# Inference in Multiply Sectioned Bayesian Networks with Extended Shafer-Shenoy and Lazy Propagation

Y. Xiang* and F.V. Jensen@
*Department of Computer Science, University of Regina
Regina, Saskatchewan, Canada S4S 0A2, yxiang@cs.uregina.ca
@Department of Computer Science, Aalborg University
DK-9220 Aalborg, Denmark, fvj@cs.auc.dk

## Abstract

As Bayesian networks are applied to larger and more complex problem domains, search for flexible modeling and more efficient inference methods is an ongoing effort. Multiply sectioned Bayesian networks (MSBNs) extend the HUGIN inference for Bayesian networks into a coherent framework for flexible modeling and distributed inference. Lazy propagation extends the Shafer-Shenoy and HUGIN inference methods with reduced space complexity.

We apply the Shafer-Shenoy and lazy propagation to inference in MSBNs. The combination of the MSBN framework and lazy propagation provides a better framework for modeling and inference in very large domains. It retains the modeling flexibility of MSBNs and reduces the runtime space complexity, allowing exact inference in much larger domains given the same computational resources.

## 1 Introduction

Bayesian networks (BNs) provide a coherent framework for inference with uncertain knowledge, and as more complex domains are being tackled, search for flexible modeling and more efficient inference methods is an ongoing effort. Multiply Sectioned Bayesian Networks (MSBNs) [11] extend the HUGIN inference method [2]. The framework allows a large domain to be modeled modularly and inference to be performed distributedly. It supports object-oriented modeling [3] and multi-agent paradigm [10]. Lazy propagation [5] extends the Shafer-Shenoy (S-S) [9] and the HUGIN methods, resulting in much reduced runtime space complexity.

We extend the lazy propagation to inference in an MSBN. The contribution is an inference scheme for MSBNs that has much reduced space complexity compared to the S-S and HUGIN-based scheme. The new scheme allows coherent inference in much larger MSBNs given the same computational resources.

We extract common aspects of tree-based inference in Section 2. We review the S-S and lazy propagation in Section 3. A distributed triangulation for MSBN compilation is presented in Section 4. We overview MSBNs in Sections 5. In Section 6, we present a new MSBN compilation. We extend the S-S and lazy propagations for inference with MSBNs in Sections 7 and 8. We compare alternative MSBN inference methods in Section 9.

We focus on the new methods without detailing most formal properties. A few necessary formal results are included with the proofs omitted due to space limit. These proofs will be included in a longer version.

## 2 Communication in trees

Consider a connected tree $T$ where each node has its (internal) state and can receive/send a message from/to a neighbor. The exchange follows the constraints:

1. Each node sends one message to each neighbor.

2. Each node can send a message to a neighbor after it has received a message from each other neighbor.

A message sent by a node is prepared on the basis of the messages received and its internal state. If the state may change as a result of messages received, then the message passing is called *dynamic* (see Fig. 1 and Section 4), otherwise called *static* (see 3.2 and 3.3). We shall refer to all the processing (outgoing message preparation and state change) taking place between receiving messages and sending a message to a particular neighbor as a generic operation called SetMsgState.

We refer to the combined activity of nodes according to the constraints as (message) *propagation*. Based on the constraints, initially only leaves can send and at any time there is a subset of nodes ready to send a message. Depending on the sending order of nodes, two regimes of propagation can be identified, *asynchronous* and *rooted*.

In asynchronous propagation, no additional rules govern the sending order. In rooted propagation, a node $r$ is arbitrarily chosen as the root, and $T$ is directed from $r$ to the leaves. All nodes except $r$ has exactly one parent. First a recursive operation CollectMessage is called in $r$. For each node $x$, when CollectMessage is called in $x$, $x$ calls CollectMessage in all children. When each child has finished with a message sent to $x$, $x$ sends a message to its parent (if any). We shall refer to this stage of rooted propagation as a (rooted) *collect* propagation.

After CollectMessage has terminated in $r$, another recursive operation DistributeMessage is called in $r$. For each node $x$ in $T$, when DistributeMessage is called in $x$, $x$ sends a message to each child and calls DistributeMessage in the child. We shall refer to this stage as a (rooted) *distribute* propagation. It is easy to show that each *asynchronous* propagation corresponds to a *rooted* propagation.

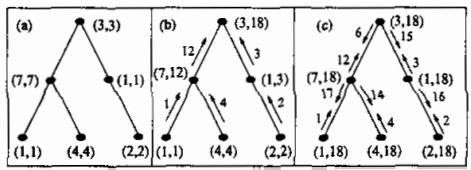

Figure 1: Dynamic propagation in a tree.

Consider Figure 1 (a). Each node stores a pair $(x, y)$, where $x$ is a local constant and $y$ is a sum initialized to $x$. To sum $x$ at all nodes, we call CollectMessage from any root (b). SetMsgState consists of adding incoming numbers to $y$, and setting the message to a neighbor $V$ as the sum of $x$ and all incoming numbers except that from $V$. The sum can now be retrieved from the root. Next, we call DistributeMessage at the same root (c). The sum can now be retrieved from any node.

## 3 Probability propagation in JTs

Various methods for inference in BNs have been constructed [6, 1, 4, 8, 9, 2]. Several [4, 9, 2] use a junction tree (JT) as runtime structure. We review how to convert a BN into a JT and then consider two of them.

### 3.1 Conversion of a BN into a JT

A BN $S$ is a triplet $(N, D, P)$ where $N$ is a set of variables, $D$ is a DAG whose nodes are labeled by elements of $N$, and $P$ is a joint probability distribution (jpd) over $N$. $D$ encodes independence in $N$ through *d-separation* [6], and hence $P(N) = \prod_{x \in N} P(x|\pi(x))$, where $\pi(x)$ is the parents of $x$ in $D$.

Conversion of a BN starts with *moralization*. It converts a DAG into an undirected graph by completing the parents of each node and dropping direction of links. The result is called a *moral* graph. Then *triangulation* (see Section 4) converts the moral graph into a *chordal* graph [7].

A JT over $N$ is a tree where each node is labeled by a subset (called a *cluster*) of $N$ and each link is labeled by the intersection (called a *sepset*) of its incident clusters, such that the intersection of any two clusters is contained in every sepset on the path between them[1].

A maximal complete set of nodes in a graph is called a *clique*. After the triangulation step, a JT for a BN is created with nodes labeled by cliques of the chordal graph. Such a JT exists iff the graph is chordal.

After a JT is created, distributions in the BN are assigned to the clusters. For each $x \in N$, $P(x|\pi(x))$ is assigned to a cluster containing $x$ and $\pi(x)$.

### 3.2 Shafer-Shenoy propagation

S-S propagation [9] is *static*, where each cluster holds a *belief table* over its variables, defined as the product of all distributions assigned to it. Hence the product of the belief tables in all clusters is the jpd.

During propagation, each message sent over a sepset is a belief table over the variables in the sepset. SetMsgState consists of multiplying the local table with incoming tables from other neighbors and marginalizing the product down to the corresponding sepset. For each cluster, after the propagation, the product of the local tables and all incoming tables is the marginal probability distribution over the variables of the cluster.

### 3.3 Lazy propagation

Lazy propagation [5] is also *static*, where each cluster $C$ holds the assigned distributions as a *set* rather than as a product. The belief table of $C$ is defined the same as above but the product is *not explicitly* computed (hence the reduced space complexity over the S-S and HUGIN methods).

Each message sent over a sepset is a set of tables each of which is over a subset of the sepset. SetMsgState to a given neighbor consists of taking the union of local tables and incoming tables from other neighbors, and then marginalizing out each variable not in the sepset.

---
[1]The property is also known as *running intersection*.

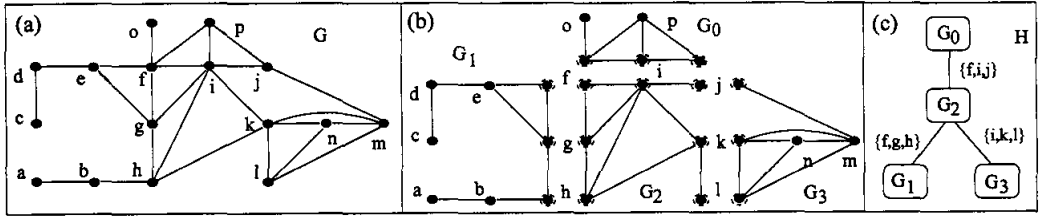

Figure 3: (a) $G$ is the union of the graphs in (b). (b) $G$ is sectioned into four subgraphs. (c) A hypertree over $G$.

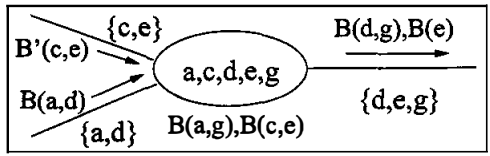

Figure 2: Message passing in lazy propagation.

Figure 2 illustrates lazy propagation. The cluster $\{a,c,d,e,g\}$ has sepsets $\{a,d\}$, $\{c,e\}$ and $\{d,e,g\}$. It has local tables $\{B(a,g), B(c,e)\}$ and receives the tables $B'(c,e)$ and $B(a,d)$. It sends out $B(d,g) = \sum_a B(a,d)B(a,g)$ and $B(e) = \sum_c B(c,e)B'(c,e)$.

## 4 Triangulation as tree propagation

We consider triangulating an undirected graph organized as a (hyper)tree.

**Definition 1** *Let $G_i = (N_i, E_i)$ ($i = 0, ..., n-1$) be $n$ graphs. The graph $G = (\cup_i N_i, \cup_i E_i)$ is the* **union** *of $G_i s$, denoted by $G = \sqcup_i G_i$.*

*If for each $i$ and $j$, $I_{ij} = N_i \cap N_j$ spans identical subgraphs in $G_i$ and $G_j$, then $G$ is* **sectioned** *into $G_i s$. $I_{ij}$ is the* **separator** *between $G_i$ and $G_j$.*

The graph in Figure 3 (a) is sectioned in (b). Each node in a separator is highlighted by a dashed circle.

**Definition 2** *Let $G = (N, E)$ be a connected graph sectioned into $\{G_i = (N_i, E_i)\}$. Let the $G_i s$ be organized as a connected tree $H$ where each node is labeled by a $G_i$ and each link is labeled by a separator such that for each $i$ and $j$, $N_i \cap N_j$ is contained in each subgraph on the path between $G_i$ and $G_j$ in $H$ [2]. Then $H$ is a* **hypertree** *over $G$. Each $G_i$ is a* **hypernode** *and each separator is a* **hyperlink***.*

Figure 3 (c) shows a hypertree $H$ over $G$ in (a). Note that the above concepts are applicable to both directed and undirected graphs.

**Definition 3** *Let $H$ be a hypertree over a graph $G$ sectioned into $\{G_i\}$. Let $G'$ be a graph from a triangulation of $G$ such that each clique in $G'$ is a subset of*

---

[2] Note the similarity to JTs.

---

*some $N_i$. Then the triangulation is* **constrained** *by $H$.*

A node $x$ in an undirected graph is *eliminated* by adding links such that all of its neighbors are pairwise linked and then removing $x$ together with links incident to $x$. The added links are called *fill-ins*.

**Theorem 4 ([7])** *A graph is chordal iff all its nodes can be eliminated one by one without adding fill-ins.*

Let a hypertree $H$ over $G$ be rooted at a given hypernode $G_i$. An elimination order $\rho$ of $G$ is *constrained* by $H$ if $\rho$ consists of recursively eliminating nodes that are only contained in a single leaf hypernode of $H$.

**Proposition 5** *An elimination order of $G$ constrained by a hypertree $H$ over $G$ produces a triangulation of $G$ constrained by $H$.*

Triangulation constrained by $H$ can be performed as a (dynamic) *rooted collect propagation of fill-ins*: Let $G_i$ be the child of $G_j$ in $H$ with separator $I_{ij} = N_i \cap N_j$. The message sent from $G_i$ to $G_j$ is a set of fill-ins over $I_{ij}$. SetMsgState consists of the following:

**Algorithm 1 (SetMsgState for propagating fill-ins)**
*add to $G_i$ fill-ins received from each neighbor except $G_j$;*
*eliminate $N_i \setminus N_j$ and add fill-ins to $G_i$;*
*set message to $G_j$ as all fill-ins over $I_{ij}$ obtained above;*

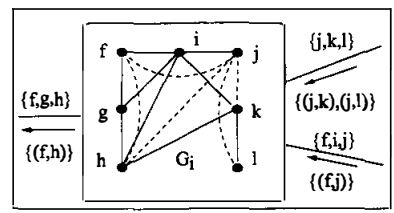

Figure 4: Hypernode $G_i$ ($i = 2$) receives fill-ins from two hyperlinks $\{f, i, j\}$ and $\{j, k, l\}$. After SetMsgState, fill-ins (dashed lines) are added to $G_i$ and the message $\{(f, h)\}$ is sent to the parent over the hyperlink $\{f, g, h\}$.

Suppose $H$ is rooted at $G_1$. For $G_1$, SetMsgState is simplified ($G_j = null$, $N_j = \phi$ and the last step is not applicable). Figure 4 illustrates the collect propagation of fill-ins.

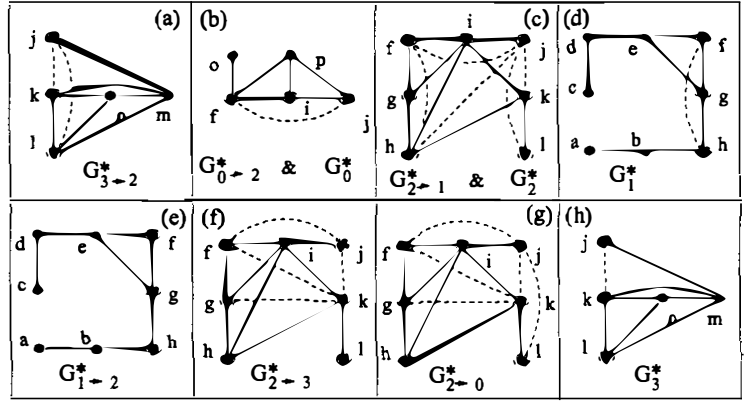

Figure 5: Illustration of propagation of fill-ins.

It can be shown that fill-ins sent during collect propagation of fill-ins is *independent* of the elimination order used by SetMsgState in each hypernode and are determined uniquely by the chosen root. Hence if $H$ has $n$ hypernodes, potentially $n$ different triangulations of $G$ (assuming each local elimination is optimized without ties) can be obtained each from a collect propagation at a distinct root. To obtain the $n$ triangulations, however, we do not have to perform collect propagation $n$ times. Instead, a full propagation in $H$ is sufficient:

CollectMessage will be performed as above. DistributeMessage will be performed with the same SetMsgState (Algorithm 1). Finally, each non-root performs SetMsgState as if it is a root.

Figure 5 illustrates the full propagation with $H$ in Figure 3. The root is $G_1$. During CollectMessage, SetMsgState is first performed in $G_3$ and $G_0$. Suppose the elimination order in $G_3$ is $(n, m)$. The fill-ins produced are $\{\{j,k\},\{j,l\}\}$ as shown in (a) with dashed links. The resultant chordal graph is labeled $G^*_{3\to 2}$. $G_3$ sends the above fill-ins to $G_2$. Similar operations then occur in $G_0$ (b) and $G_2$ (c).

Since $G_1$ is the root, it performs a simplified SetMsgState. After adding the fill-in $\{f, h\}$, the resultant graph $G^*_1$ is chordal as shown in (d). CollectMessage now terminates. DistributeMessage follows as shown in (e) to (g). Each non-root hypernode performs one more SetMsgState as if it is a root with the results shown in (b), (c) and (h). Note that in (h), since the received fill-in is $\{j,k\}$ and the elimination can be performed in any order, $G^*_3$ is simpler than $G^*_{3\to 2}$.

## 5  Overview of MSBNs

An MSBN $M$ is a collection of Bayesian subnets that together defines a BN [11, 10]. $M$ represents probabilistic dependence of a *total universe* partitioned into multiple *subdomains* each of which is represented by a subnet.

Just as the structure of a BN is a DAG, the structure of an MSBN is a multiply sectioned DAG (MSDAG) with a hypertree organization:

**Definition 6** A hypertree MSDAG $\mathcal{D} = \bigsqcup_i D_i$, where each $D_i$ is a DAG, is a connected DAG such that (1) there exists a hypertree over $\mathcal{D}$, and (2) each hyperlink d-separates [6] the two subtrees that it connects.

The second condition requires that nodes shared by two subnets form a *d-sepset*:

**Definition 7** Let $D_i = (N_i, E_i)$ $(i = 0, 1)$ be two DAGs such that $D = D_0 \sqcup D_1$ is a DAG. The intersection $I = N_0 \cap N_1$ is a d-sepset for $D_0$ and $D_1$ if for every $x \in I$ with its parents $\pi(x)$ in $D$, either $\pi(x) \subseteq N_0$ or $\pi(x) \subseteq N_1$. Each $x \in I$ is called a d-sepnode.

This is established as follows:

**Proposition 8** Let $D_i = (N_i, E_i)$ $(i = 0, 1)$ be two DAGs such that $D = D_0 \sqcup D_1$ is a DAG. $N_0 \setminus N_1$ and $N_1 \setminus N_0$ are d-separated by $I = N_0 \cap N_1$ iff $I$ is a d-sepset.

It can be shown that the above definition of MSDAG is equivalent to the constructive definition in [11]. An MSBN is defined as follows:

**Definition 9** An MSBN $M$ is a triplet $M = (\mathcal{N}, \mathcal{D}, \mathcal{P})$. $\mathcal{N} = \bigcup_i N_i$ is the total universe where each $N_i$ is a set of variables. $\mathcal{D} = \bigsqcup_i D_i$ (a hypertree MSDAG) is the structure where nodes of each DAG $D_i$ are labeled by elements of $N_i$. Let $x$ be a variable and $\pi(x)$ be all parents of $x$ in $\mathcal{D}$. For each $x$, exactly one of its occurrences (in a $D_i$ containing $\{x\} \cup \pi(x)$) is assigned $P(x|\pi(x))$, and each occurrence in other

*DAGs is assigned a constant table.* $\mathcal{P} = \prod_i P_{D_i}$ *is the* **jpd**, *where each* $P_{D_i}$ *is the product of the probability tables associated with nodes in* $D_i$. *A triplet* $S_i = (N_i, D_i, P_{D_i})$ *is called a* **subnet** *of* $M$.

An example MSBN is shown in Figure 6.

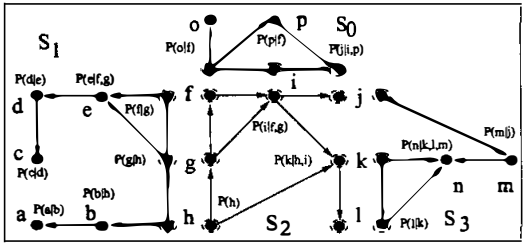

Figure 6: An MSBN.

## 6 Compilation of MSBNs

So far, inference in MSBNs [11, 10] has been an extension to the HUGIN method [2][3], which works with one triangulation and one decomposition of messages for the entire propagation. As demonstrated in Section 4 and below, it is possible to let the triangulation and decomposition depend on the direction of messages. The resultant clusters can be smaller than obtained by the HUGIN method. Below we explore this idea for inference in MSBNs using the S-S and lazy propagation.

### 6.1 Local structure for message/inference

First *moralization* is performed as a full dynamic propagation on the hypertree. A message sent from a hypernode to another consists of (moral) links over their d-sepset. During CollectMessage, SetMsgState consists of the following: (1) For each hypernode, parents of each node in $D_i$ are completed and directions of links are dropped. (2) Moral links from each child hypernode are then added. (3) Set the message to the parent hypernode as the moral links over their d-sepset. For DistributeMessage, SetMsgState consists of (2) and (3). Figure 3 (b) is the moralization of the MSBN in Figure 6.

Next *triangulation* is performed as in Section 4. Then we convert each $G_i^*$ into a JT for local inference (as in Section 3.1) and convert each $G_{i \rightarrow j}^*$ into a junction forest (JF) for computing messages from subnet $S_i$ to $S_j$ for inter-subnet belief propagation. We present the conversion of $G_{i \rightarrow j}^*$ into a message JF below:

To see the need of multiple structures for each subnet, observe that $G_i^*$ is generally more densely connected than $G_{i \rightarrow j}^*$. In Figure 5, the d-sepset is complete in $G_1^*$, but incomplete in $G_{1 \rightarrow 2}^*$. By using $G_{i \rightarrow j}^*$, the message from $S_1$ to $S_2$ can be decomposed into two submessages, one over $\{f, g\}$ and the other over $\{g, h\}$. This results in a more compact message representation. For each $G_{i \rightarrow j}^*$, we organize its cliques into a set of JTs (a JF) so that each submessage can be obtained *directly* from one cluster of each JT. Without formally presenting the general algorithm, we illustrate using the example in Figure 5.

---
[3]The HUGIN propagation is *dynamic* whereas S-S as well as lazy propagation are *static*.

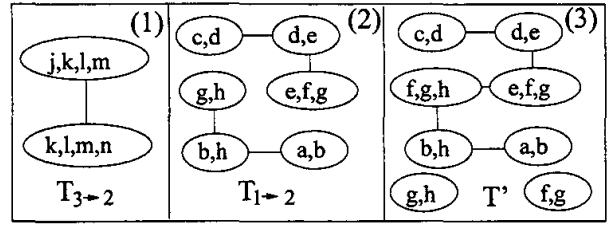

Figure 7: Junction forests for message computation.

First, consider $G_{3 \rightarrow 2}^*$. Since the d-sepset is complete (no opportunity for message decomposition), we organize the cliques of $G_{3 \rightarrow 2}^*$ into a JT $T_{3 \rightarrow 2}$ shown in Figure 7 (1). During inference, the message from $S_3$ to $S_2$ can then be obtained from the cluster $\{j, k, l, m\}$. Similarly, JTs $T_{0 \rightarrow 2}$, $T_{2 \rightarrow 1}$ and $T_{2 \rightarrow 0}$ can be obtained.

Next, consider $G_{1 \rightarrow 2}^*$. Since the d-sepset is incomplete (the message is decomposable), we create a JF consisting of two JTs as in (2). During inference, the submessage over $\{f, g\}$ can be computed using the upper JT from the cluster $\{e, f, g\}$. The submessage over $\{g, h\}$ can be obtained from the cluster $\{g, h\}$ of the lower JT.

The JF is constructed as follows: For each clique in the subgraph of $G_{1 \rightarrow 2}^*$ spanned by the d-sepset, create an isolated node labeled by the clique. Hence we obtain the two clusters at the bottom of (3). They are the candidate clusters from which the submessages will be obtained. We then complete the d-sepset in $G_{1 \rightarrow 2}^*$ and create a JT out of it as shown in the top of (3). We split this JT into two and merge each with one of the candidate clusters as follows:

We delete the d-sepset cluster $\{f, g, h\}$, breaking the JT into two subtrees. For one subtree, the cluster $\{b, h\}$ was adjacent to $\{f, g, h\}$. Since the candidate cluster $\{g, h\}$ satisfies $\{g, h\} \cap \{b, h\} = \{f, g, h\} \cap \{b, h\}$, we connect $\{g, h\}$ with $\{b, h\}$. For the other subtree, the cluster $\{e, f, g\}$ was adjacent to $\{f, g, h\}$. Since the candidate cluster $\{f, g\}$ is a subset of $\{e, f, g\}$, we remove the candidate cluster $\{f, g\}$. The resultant JF is the one in (2). Similarly, JF $T_{2 \rightarrow 3}$ can be obtained.

Without confusion, we refer to message JFs and inference JTs collectively as JFs. In the next section, we define a data structure to guide message passing between local JFs at adjacent subnets.

### 6.2 Linking message JFs and inference JTs

Inference in an MSBN can be performed as a full propagation in the hypertree consisting of message passing among JFs (SetMsgState will be detailed later). When a message is to be sent from $S_i$ to $S_j$, it is computed using $T_{i \to j}$. When $S_j$ receives the message, it will be processed by $T_j$ and each $T_{j \to k}$ ($k \neq i$). Figure 8 (1) illustrates directions of messages during collect propagation with root $S_1$, and (2) illustrates distribute propagation.

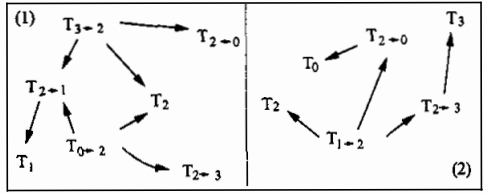

Figure 8: Directions of messages during propagation.

As each submessage is obtained from a cluster of the sending JF and absorbed into a cluster of the receiving JF, we create a *linkage* that links the pair of clusters.

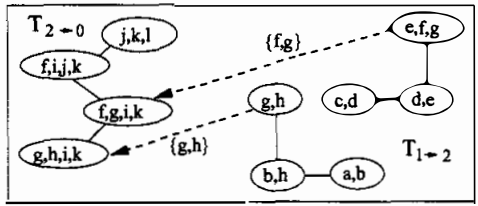

Figure 9: Linkages between two message JFs.

Figure 9 shows the two linkages from $T_{1 \to 2}$ to $T_{2 \to 0}$ used during distribute propagation. It reflects the fact that the d-sepset $\{f, g, h\}$ can be decomposed into two independent subsets $\{f, g\}$ and $\{g, h\}$ conditioned on their intersection $\{g\}$. Each linkage (shown as a dashed arc) is labeled by the intersection of the two end clusters. We shall call the two clusters the *hosts* of the linkage. Once linkages are determined, the set of all JFs forms a *linked junction forest* (LJF).

### 6.3 Belief assignment

Next, we assign conditional probability tables (CPTs) in the MSBN to clusters in the LJF. For each JF of each subnet, the assignment is performed as follows: For each variable $x$, if a CPT is associated with it, then assign the CPT to a cluster in the JF that contains $x$ and its parents.

The *joint system belief* of the LJF is then defined as $B(\mathcal{N}) = \prod_i \prod_j \prod_k \beta_{i,j,k}$, where $i$ is the index of inference JTs, $j$ is the index of clusters in a given JT, $\beta_{i,j}$ denotes the set of CPTs assigned to the $j$th cluster in the $i$th JT, and $\beta_{i,j,k}$ is the $k$th CPT in the set. It is easy to see that $B(\mathcal{N})$ is identical to the jpd of the MSBN.

Since CPTs are assigned in the same way in inference JTs and message JFs, the belief of all JFs from the same subnet are identical.

Although each subnet is associated with multiple JFs, only one copy of each CPT needs to be physically stored. For each CPT, it suffices to store a pointer at the assigned cluster in each JF.

## 7 Shafer-Shenoy propagation in LJF

We extend the S-S propagation (Section 3.2) for inference in a linked junction forest.

For each cluster in each JF of each subnet, a belief table is created by multiplying the CPTs assigned to the cluster. Inference is performed as a full propagation over the hypertree during which messages are sent between JFs in adjacent subnets. When a message JF has multiple linkages to an adjacent JF, the message consists of multiple submessages (otherwise the message consists of a single submessage) each of which is sent across a distinct linkage. Each linkage is used for message passing in a unique direction.

Each submessage is prepared at a distinct JT in a message JF. A local collect S-S propagation is started at the linkage host and the submessage is then obtained at the host. The propagation involves incoming linkages and their hosts in the adjacent JFs, as illustrated in Figure 10.

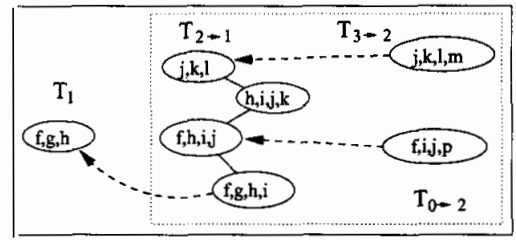

Figure 10: To compute the submessage from $T_{2 \to 1}$ to $T_1$, $T_{2 \to 1}$ is extended (dotted box) to include linkage hosts $\{j, k, l, m\}$ from $T_{3 \to 2}$ and $\{f, i, j, p\}$ from $T_{0 \to 2}$. The collect propagation starts at linkage host $\{f, g, h, i\}$.

Now we define SetMsgState for preparing the message from $S_i$ to $S_j$ sent by message JF $T_{i \to j}$:

**Algorithm 2 (SetMsgState for S-S propagation in LJF)**

*for each junction tree of $T_{i \to j}$*
  *start collect S-S propagation at the host of linkage to $S_j$;*
  *set submessage as marginal of host belief to the linkage;*

To analyze the effect of the propagation, we define the belief tables associated with different identities in an LJF: For each cluster $C$ with a local belief table $\beta$ and incoming messages $\beta_i$ ($i = 1, ...$), the belief table $B_C(C)$ is the product $\beta * \prod_i \beta_i$. Note that the messages include messages from sepsets as well as submessages from linkages. For each inference JT $T$ over $N$, the belief table $B_T(N)$ is the product $B_T(N) = \prod_i B_{C_i}(C_i)$, where $i$ indexes clusters of $T$. It can be shown that the extended S-S propagation is coherent.

After the extended S-S propagation in the LJF, a S-S propagation needs be performed at an inference JT to answer local queries. Note that the collect stage of the propagation should be performed on the *extended* JT to count the incoming messages from adjacent message JFs. Also note that when evidence is available on a variable in a subnet, it should be entered to a relevant cluster in *each* JF of the subnet.

## 8 Lazy propagation in LJF

The extended S-S propagation can be directly modified into *extended lazy propagation* in LJFs as follows:

For each cluster in each JF of each subnet, its belief table is *defined* in the same way as the extended S-S propagation, but multiplication of assigned CPTs is *not* performed explicitly. The S-S propagation performed in each JF is replaced by lazy propagation (Section 3.3). Each message over a sepset and each submessage over a linkage will in general be a set of belief tables over a subset of variables of the sepset/linkage *without* being multiplied together. Theorem 10 shows that the extended lazy propagation ensures coherent inference.

**Theorem 10** *After a full extended lazy propagation in an LJF, for each subnet $S_i$ over $N_i$, its inference JT $T_i$ satisfies $B_{T_i}(N_i) = \sum_{N \setminus N_i} \prod_j B_{T_j}(N_j)$, where $j$ indexes inference JTs.*

As for normal BNs, the main advantage of lazy propagation is its decomposed representation of belief tables/messages. The decomposition leads to reduced space complexity, which is particularly significant when the problem domain is very large.

## 9 Conclusion

We presented how to construct a linked junction forest (LJF) from a multiply sectioned Bayesian network (MSBN), and how to extend Shafer-Shenoy and lazy propagation for inference in such an LJF. It is worthwhile to compare the new methods with earlier work on the construction of LJF and the HUGIN based inference method [11, 10].

First of all, the new method constructs multiple JFs for each subnet, one for local inference and the others for inter-subnet message computation. The previous method, on the other hand, creates a single JT at each subnet for both local inference and inter-subnet message computation. With the new method, since each message JF is dedicated to the computation of messages to a particular subnet, its structure is less constrained and is generally more sparse. With the previous method, a JT must function correctly at all conditions (send and absorb messages to/from each adjacent subnet) and it is thus more constrained, resulting in generally more densely connected JT structures.

Although we have extended the S-S and lazy propagations in the LJF constructed by the new method, they can be modified to perform in an LJF constructed by the previous method as well. Given what we have presented, the modification is straightforward. To the S-S propagation, the benefit of using the new construction is more compact belief representation and more efficient inference computation due directly to the sparser JF structure. To lazy propagation, the benefit is that the sparser structures provide better guidance to the propagation. To see this, imagine that if an entire message JF is a single cluster, the burden of finding an effective marginalization order for computing a message will be placed entirely at runtime. Hence, each message JF in the new construction can be viewed as a concise recording of a set of effective marginalization orders ready for runtime exploitation.[4] On the other hand, the LJF by the previous method needs not to maintain multiple JFs at each subnet. Inter-subnet message computation and local inference computation can then be completed by just one propagation in the only JT at a subnet (instead of several propagations one at each JF).

This observation suggests a tradeoff between using an LJF constructed by the new method and that by the previous method. One factor in making a choice is the relative sparseness of the LJF obtained by each method, which depends on the topology of the MSBN in question. Another factor in practice is the emphasis placed on simplicity in control (which translates to development time) and efficiency in runtime computation.

Secondly, the extended lazy propagation has much lower space complexity than the previous HUGIN based inference for MSBNs due to the factorized storage of belief. With the lazy propagation, for each CPT in the MSBN, only one copy needs to be stored in the LJF. Hence the total number of independent parameters stored in the LJF is 46 for the example MSBN. If full CPTs are stored to save the on-line derivation,

---

[4]A marginalization order specifies the order in which each variable is to be marginalized out. Two such orders are *equally* effective if their computational complexity are the same.

92 values should be stored. With the HUGIN based method, the total storage of all belief tables for all clusters in the sparsest LJF has a size of 140. As the MSBN grows in size and connectivity, the sizes of clusters of the LJF grow. The belief storage per cluster in an LJF grows exponentially with the cluster size with the HUGIN based method, while with the extended lazy propagation it grows only linearly. Therefore, the extended lazy propagation will allow much larger MSBNs to be constructed and used than possible with the HUGIN based inference, given one's computational resource.

## Acknowledgement

We thank the anonymous reviewers for helpful comments. The support of Research Grant OGP0155425 to the first author from NSERC of Canada is acknowledged.